\documentclass[11pt,a4paper]{article}
\usepackage[hyperref]{acl2020}
\usepackage{times}
\usepackage{graphicx}
\usepackage{multirow}
\usepackage{latexsym}

\usepackage{microtype}

\aclfinalcopy 

\title{An Experimental Study of The Effects of Position Bias on Emotion Cause Extraction}

\author{Jiayuan Ding \\
  VMware \\
  Palo Alto, CA 94304 \\
  \texttt{jiayuand@usc.edu} \\\And
  Mayank Kejriwal \\
  Information Sciences Institute \\
  University of Southern California \\
  Los Angeles, CA 90292 \\
  \texttt{kejriwal@isi.edu} \\}

\date{}

\begin{document}
\maketitle
\begin{abstract}
  Emotion Cause Extraction (ECE) aims to identify emotion causes from a document after annotating the emotion keywords. Some baselines have been proposed to address this problem, such as rule-based, commonsense based and machine learning methods. We show, however, that a simple random selection approach toward ECE that does not require observing the text achieves similar performance compared to the baselines. We utilized only position information relative to the emotion cause to accomplish this goal. Since position information alone without observing the text resulted in higher F-measure, we therefore uncovered a bias in the ECE single genre Sina-news benchmark. Further analysis showed that an imbalance of emotional cause location exists in the benchmark, with a majority of cause clauses immediately preceding the central emotion clause. We examine the bias from a linguistic perspective, and show that high accuracy rate of current state-of-art deep learning models that utilize location information is only evident in datasets that contain such position biases. The accuracy drastically reduced when a dataset with balanced location distribution is introduced. We therefore conclude that it is the innate bias in this benchmark that caused high accuracy rate of these deep learning models in ECE. We hope that the case study in this paper presents both a cautionary lesson, as well as a template for further studies, in interpreting the superior fit of deep learning models without checking for bias.
\end{abstract}

\section{Introduction}
Emotion cause extraction (ECE), a sub-task of emotional analysis, aims to identify the potential causes behind a certain emotion expression in the document. Due to ECE's potentially widespread applications \cite{Lee:10,Russo:11,Gao:15,Chengchen:17}, it has attracted a lot of attention from the academic community in recent years. In comparison with emotion classification  \cite{Pang:02,Gao:13,Ding:16}, ECE is more challenging because it requires a deeper understanding of the semantics of the text before it can accurately extract the emotion causes. The task is initially defined as a word-level sequence labeling problem in \citet{SophiaLee:10}; however, the emotion causes often appear at the clause-level. Therefore, \citet{Gui:16a} released a new corpus and re-formalized the ECE task as a clause-level classification problem. This corpus has received much attention after its release and has become a standard benchmark for ECE research.

Figure 1 provides an example instance in the benchmark. There are four clauses in this instance. Clause $c_{4}$ is the emotion clause that contains the emotion keyword ``touched", and the clause $c_{3}$ describes the corresponding cause. The goal of ECE is to predict, for each clause in the instance, whether the clause is an emotion cause, given the annotation of the emotion expression.

\begin{figure}[htb] 

\setlength{\abovecaptionskip}{0pt}
\setlength{\belowcaptionskip}{0pt}
\centering
\includegraphics[width=3.3in]{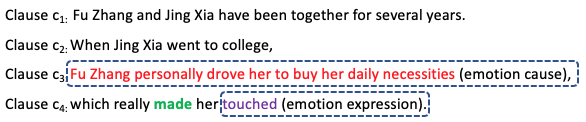}
\caption{An instance of the ECE task as exemplified from the benchmark} 
\end{figure}

In the past, several techniques have been proposed to address this problem, including rule based methods \cite{Lee:10,Lee:13,SophiaLee:10,Gui:16a} and traditional machine learning based methods\cite{Lee:10,Gui:14,Russo:11}. These methods are primarily based on linguistic rules or lexicon features.

More recently, some deep neural networks have been applied to solve this task and shown to outperform previously proposed approaches. It was claimed that these newer models were able to get a better understanding of the document, which led to their improved performance. For example, \citet{RTHN} suggested that high performance of their model is contributed mainly through learning the correlation between multiple clauses in a document, and the correlation includes two key features. One of them is that two clauses with similar semantics are equally likely to be the emotion cause. The other attribute is that content information of clauses around the target clause could help to infer whether if it is the emotion cause. 

Similarly, \citet{PAEDGL} believed that their deep neural network model benefited greatly from encoding three elements (text content, relative position and global label). In \citet{HCS}, a hierarchical network-based framework was proposed to solve this task via extracting features, including word's position, semantic levels (word and phrase) and interaction among clauses. In addition, \citet{CANN} proposed the Co-Attention Neural Network method, which involves comparing only one candidate clause at a time with the annotated emotion, to see if it indeed describes the cause without the aid of other candidate clauses. 

In this paper, we empirically prove that it is not the deep semantic information as claimed by the above deep neural network models, but the position information of emotion cause that \emph{primarily} contributes to high accuracy. In fact, the position feature used by these models is an inherent bias in the ECE benchmark itself. 
Our experiments, which controlled for this `position skew' without changing the text itself, showed that these models' performance decreased significantly by an average of 18 percent.

The main contributions of this work can be summarized as follows:

\begin{itemize}
\item[1.] We show a deep position bias for ECE in the current benchmark and argue that it is not sufficiently demanding for the task. Specifically, since the cause clause immediately precedes the emotion clause in so many instances, complex models achieve good performance without using the actual text context.  
\item[2.] We study and explain the existence of the bias in the benchmark from a linguistic perspective. Since all instances in this benchmark belong to SINA News, the unique and uniform diction and syntax of this genre make it even easier to accomplish the ECE task. 
\item[3.] Current high performance of deep neural network models make use of the bias existing in the benchmark to achieve better performance. We show that, after controlling for the bias in the benchmark, the performance of these models decreased drastically. 
\item[4.] Finally, we improve the benchmark by debiasing it. We prove the efficacy of our debiasing by comparing the accuracy of a simple random approach that we devised to three deep neural network models on the debiased benchmark. We verify that this change makes the new benchmark adequately challenging for the purpose of testing the efficacy of these new models. 
\end{itemize}

\section{Related Work}
The task on emotion cause extraction was first presented by \citet{Lee:10}. They manually constructed a small-scale corpus from Academia Sinica Balanced Chinese Corpus. The ECE task here was formalized as a word-level sequence labeling problem. Rule-based or machine learning methods were then proposed based on manually designed rules or features. Two sets of linguistic features with the help of linguistic cues were developed to detect emotion causes by \citet{Lee:10}. A similar rule-based method was also proposed by \citet{Lee:13}. 

One emotion cause corpus from Chinese microblog posts was constructed by \citet{LiXu-2014} and a rule-based approach was proposed to identify emotion cause by importing knowledge from the field of sociology. Another emotion cause corpus consisting of annotations for 1,333 microblogs was also constructed by \citet{Gui:14}. Based on this corpus, 25 rules were manually developed, and two machine learning based methods, SVMs and conditional random fields (CRFs), were proposed to infer emotion causes. Using also this dataset, \citet{Gao-2015a} and \citet{Gao-2015b} proposed a rule-based approach, which extracted various linguistic cues from the annotated corpus and generalized linguistic rules with the help of these cues. Other studies \cite{Russo:11,Neviarouskaya-2013,Ghazi-2015,Song-2015,Yada-2017}, which developed their own corpus to tackle the ECE task, also came up with various rule-based or machine learning methods.

Because clause works better as a unit of emotion cause when compared to phrase or word, \citet{Gui:16a} and \citet{Gui-16b} released a Chinese emotion cause corpus from SINA city news and re-formalized the ECE task as a clause-level classification problem. Given the annotation of emotion expression, the goal is to predict, for each clause in the document, whether it is an emotion cause. This corpus has received much attention from the studies thereafter and has become a standard benchmark for ECE research. Rule-based or traditional machine learning approaches \cite{Gui:16a,Gui-16b,Xu-2017} were applied on this benchmark. In the last three years, some deep neural network models were also brought up to target this benchmark. For example, a hierarchical architecture based on RNN and Transformer was developed by \citet{RTHN}. The lower layer of the architecture is a word-level encoder consisting of multiple RNNs to obtain clause representation. The upper layer is a clause-level encoder based on a stacked Transformer, where the clause representation is repeatedly learned and updated by incorporating relations among multiple clauses. \citet{PAEDGL} proposed an encoder-decoder style method to incorporate relative position and global label information into clause representation. 

Our work differs from all of the above significantly in that we are not proposing a new algorithm for ECE. We are going to uncover a position bias of emotion cause clause in the ECE benchmark and prove that current high performance deep neural network models make use of this loophole. After alleviating the bias in the benchmark, the performance of these models decrease drastically.

\section{Position Bias of Emotion Causes in the ECE Benchmark}
\subsection{Details of the Benchmark}
\citet{Gui:16a} and \citet{Gui-16b} released Emotion Cause Extraction (ECE) corpus based on Chinese SINA city news. Blog posts over a 3 year period (2013-15) from Chinese city news \footnote{http://news.sina.com.cn/society/} containing 20,000 articles were filtered and selected as the raw corpus. After referencing a dictionary of 10,259 Chinese emotion keywords \cite{Xu-2008}, 15,687 emotion instances were extracted by keyword matching. For each matched keyword, the authors extracted a few preceding and following clauses as the context. However, the presence of keywords does not necessarily convey important emotional information. As a result, \citet{Gui:16a} and \citet{Gui-16b} manually removed irrelevant instances, which resulted in 2,105 instances.

\begin{table}[ht]
\setlength{\abovecaptionskip}{0pt}
\setlength{\belowcaptionskip}{0pt}
\begin{center}
\scalebox{0.90}{
\begin{tabular}{|l|rl|}
\hline \bf Item & \bf Number & \\ \hline
\hline
Instance & 2,105 & \\
Emotion Cause & 2,167 & \\
\bf Documents with 1 emotion cause & \bf 2,046 & \\
Documents with 2 emotion causes & 56 & \\
Documents with 3 emotion cause & 3  & \\
\hline
\end{tabular}
}
\end{center}
\caption{ Some details of the benchmark. The three document sets (with 1, 2 and 3 emotion causes) add to 2,105 (the total number of instances).}
\label{tab:stats}
\end{table}

\begin{table}[ht]
\setlength{\abovecaptionskip}{0pt}
\setlength{\belowcaptionskip}{0pt}
\begin{center}
\scalebox{0.80}{
\begin{tabular}{|l|rl|}
\hline \bf Position & \bf Percentage & \\ \hline
\hline
Previous 10 Clauses &  0.04\% &  \\
Previous 9 Clauses &  0.04\% &  \\
Previous 7 Clauses &  0.13\% & \\
Previous 6 Clauses &  0.32\% & \\
Previous 5 Clauses &  0.32\% & \\
Previous 4 Clauses &  0.59\% & \\
Previous 3 Clauses &  1.70\% & \\
Previous 2 Clauses &  8.12\% & \\
\bf Previous 1 Clauses & \bf 54.45\% &  \\
\bf In the same clauses & \bf 23.58\% &  \\
Next 1 Clauses &  7.47\% & \\
Next 2 Clauses &  2.21\% & \\
Next 3 Clauses &  0.50\% & \\
Next 4 Clauses &  0.18\% & \\
Next 5 Clauses &  0.09\% & \\
Next 7 Clauses &  0.04\% & \\
Next 8 Clauses &  0.09\% & \\
Next 12 Clauses &  0.04\% & \\
\hline
\end{tabular}
}
\end{center}
\caption{ Position distribution of cause clauses relative to emotion clause in the original benchmark. The vast majority (about 78\%) of the causes occur in the same, or immediately preceding, clause as the emotion clause.}
\label{tab:position}
\end{table}

Each instance in the dataset contains only one emotion keyword and at least one emotion cause. The numbers of extracted instances and emotion cause are both listed in Table \ref{tab:stats}. 97.2\% of the instances have only one emotion cause while others have two or three emotion causes. An example with two emotion causes is that ``A policeman gave the lost money back to the old man. The police also told him that the thief was caught and sent to jail. The old man was very happy." In this example, there are two causes for happiness. One is that he got money back, and the other is that thief was punished. 
Table \ref{tab:position} shows the distribution of cause clause position relative to the emotion clause. On the surface, the data makes it evident that a position of `0' or `-1' of the cause clause relative to the emotion clause is frequent (78\% of instances) and  repetitive. We suspect this `position skew' did not gain the necessary attention from earlier works although it was mentioned before.

\subsection{Emotion Cause Detection Without Observing the Context}

\subsubsection{Methodology}
Our experimental study continues to regard ECE as a clause-level sequence labeling problem. However, since 97.2\% of the instances have a single emotion cause, we chose to select one clause with the highest probability to be the emotion cause in each instance, instead of predicting if every clause could be an emotion cause. To support the experimental study, we devised a random selection approach that relies on the overall probability distribution of each possible position (designated \textbf{Random}) to tackle the ECE task, which does not involve understanding the context (i.e. looking at the text). Using the statistical distribution of cause clause locations listed in Table \ref{tab:position} as our sampling distribution, we utilized the numpy.random.choice() function provided by Python NumPy package to randomly sample one clause per instance as the cause of the emotion in that instance. 

\subsubsection{Experimental Setting}
We followed the same experimental setup as \citet{Gui:16a,Gui-16b}. 90\% of the dataset was stochastically selected as training data while the other 10\% was chosen as testing data. This method was evaluated 25 times with different train/test splits.

For evaluation metrics, we also followed the same metrics as \citet{Gui:16a,Gui-16b}. The precision, recall and F score were used as the metrics for evaluation. These metrics in emotion cause extraction are defined by\\

Precision =  $\frac{\sum correct\_causes}{\sum proposed\_causes} $\\

Recall =  $\frac{\sum correct\_causes}{\sum annotated\_causes} $\\

F\-measure =  $\frac{2 * Precision * Recall}{Precision + Recall} $\\

In these formulas, \textsl{proposed\_cause} denotes the number of clauses predicted to be emotion causes; \textsl{annotated\_cause} denotes the number of clauses actually labeled as emotion causes in the  ground truth; \textsl{correct\_cause} denotes the number of clauses labeled as the emotion cause \emph{and} predicted correctly by the model as the emotion cause (i.e. the true positives).

\subsubsection{Evaluation and Comparison}
We compared our approach with the following baseline methods:

\begin{itemize}
\item \textbf{RB} (Rule based method): The method based on manually defined linguistic rules, as proposed by \citet{Lee:10};
\item \textbf{CB} (Common-sense based method): The method based on commonsense knowledge, as proposed by \citet{Russo:11};
\item \textbf{RB+CB} (Rule based + Common-sense based method): This method is a combination of RB and CB;
\item \textbf{RB+CB+ML} (Machine learning method trained from rule-based features and common-sense knowledge base): This method, proposed by \citet{Lee:10}, utilizes linguistic rules and facts in a knowledge base as features for classification. 
\end{itemize}

\begin{table}[ht]
\setlength{\abovecaptionskip}{0pt}
\setlength{\belowcaptionskip}{0pt}
\begin{center}
\scalebox{0.90}{
\begin{tabular}{|l | l | l | rl |}
\hline \bf Method & \bf P & \bf R & \bf F1 & \\ \hline
\hline
RB & 0.6747 & 0.4287  & 0.5243 & \\ 
CB & 0.2672  & 0.7130  & 0.3887 & \\ 
RB+CB & 0.5435  & 0.5307  & 0.5370 & \\ 
RB+CB+ML & 0.5921  & 0.5307  & 0.5597 & \\ \hline
Random & 0.5512 & 0.5359 & 0.5434 & \\ 
\hline
\end{tabular}
}
\end{center}
\caption{ Comparing Precision, Recall and F1 of different baseline methods on  the original benchmark.}\label{tab:performance}
\end{table}

The performance results are shown in Table \ref{tab:performance}. Even without observing the context, our approach achieved a result that is almost indistinguishable (< 2\%) from the best baseline on F1. The high performance of our proposed method was contributed solely by our knowledge of the relative positions of emotion cause clause, in which the instances of ``-1'' position occupy a majority (54.45\%). This result illustrates the bias in the benchmark. Many models, especially deep neural network models, can achieve high performance because of this position bias and not the ability of learning deep semantics from texts, as originally hypothesized. We revisit this idea in Section 5, where some deep neural network models will be re-evaluated on a modified version of this benchmark that has a different, more balanced distribution of cause clauses in terms of their position relative to the emotion clause.

\section{The Linguistic Origin of Position Bias}

Because of the high occurrence of emotion causes located at ``-1'' and ``0'' position, we will focus mainly on analyzing these two types of conditions. Based on the linguistic analysis, \citet{Lee:10} identified groups of linguistic cue words that were highly correlated with emotion causes. In this paper, we manually group and count the number of each set of linguistic cue words that appear in ``-1'' and ``0'' location, as shown in Tables 4 and 5 respectively. Each group of the linguistic cues serves as an indicator that marks the emotion cause.

Group I in Table 4 lists prepositions that precede the stated reason behind an emotional reaction specified in the next clause while in Table 5, the prepositions are used immediately after emotion keywords to connect with the reason that is to follow in the same clause. Group II in Tables 4 and 5 lists the conjunctions that are used to directly convey cause and effect relationships. Group III in both Tables 4 and 5 includes three common light verbs, which have little semantic content of their own but are often used to directly address how something makes one feel emotionally. Group IV and V in Table 4, and Group IV in Table 5, comprise words that are often used in news reports under the context of having protagonists (or affected ones) to recall what has happened. Group V in Table 5 represents a specific syntax format that conveys how something makes someone feel in the passive tense. 

The five groups of cue words in Table 4 can explain 51.19\% of ``-1'' instances, and the five groups of cue words in Table 5 can explain 86.69\% of ``0'' instances. In other words, the majority of ``-1''and ``0'' occurrences can be explained from a linguistic perspective. More specifically, the high frequency of reported verbs and epistemic verbs, which are characterizing diction of Chinese internet news report, shows that bias in this benchmark is due to the use of only a single genre of literature that has a monolithic, standardized writing style. Thus, position bias is not the only issue in using this benchmark to assess the general state of ECE systems.

\begin{table*}[htbp]
\setlength{\abovecaptionskip}{0pt}
\setlength{\belowcaptionskip}{0pt}
\begin{center}
\scalebox{0.70}{
\begin{tabular}{|l | p{5cm} | p{9cm} | rl |}
\hline \bf Group & \bf Cue Words & \bf Examples & \bf Percentage & \\ \hline
\hline
I: Prepositions  & `{\bf for}': dui4, dui4yu2, dui4ci & \textbf{For} his son's irrational behavior, Gang Luo became very angry. &  117/1180 = \textbf{9.92\%} & \\ \hline

II: Conjunctions  & {`{\bf because}': yin1, yin1wei4,\newline you2yu2 \newline \newline `{\bf so}': yu1shi4, suo3yi3, yin1er2} & \textbf{Because} Yong Jiang has adoptive parents who are open-minded, he is therefore happier. &  85/1180 = \textbf{7.20\%} & \\ \hline

III: Light Verbs  & `{\bf make}': rang4, ling4, shi3 & Fuhua Zhang personally drove her to buy her quilt and other daily necessities, which really \textbf{makes} her touched. &  151/1180 = \textbf{12.8\%} & \\ \hline

IV: Reported Verbs  & `{\bf to think about}': xiang3dao4,\newline xiang3qi3, hui2xiang3qi3, \newline yi1xiang3, xaing3lai3 \newline \newline
`{\bf to talk about}': shuo1dao4, \newline shuo1qi4, yi1shuo1, jiang3dao4, \newline jiang3qi3, yi1jiang3, tan2dao4, \newline tan2qi3, yi1tan2, ti22dao4, \newline ti2qi3, yi1ti2 \newline \newline
`{\bf to remember of}': hui2yi4, \newline hui2yi4qi3
 & When \textbf{mentioning} the situation at the time, Mr. Du still had a lingering fear. &  70/1180 = \textbf{5.93\%} & \\ \hline

V: Epistemic Markers  & `{\bf to hear}': ting1, ting1dao4, \newline ting1shuo1 \newline \newline
`{\bf to see}': kan4, kan4dao4, \newline kan4jian4, jian4dao4, \newline jian4, kan4dao4, \newline kan4jian4 \newline \newline
`{\bf to know}': zhi1dao4, \newline de2zhi1, de2xi1,huo4zhi1, \newline huo4xi1, fa1xian4, fa1jue2
 & After \textbf{hearing} about the other party's death, it is difficult for her to suppress the grief in her heart. &  181/1180 = \textbf{15.34\% }& \\ \hline

  &  &  & \bf 51.19\% & \\ \hline

\end{tabular}
}
\end{center}
\caption{ Linguistic cue words that characterize emotion cause at ``-1'' position. For each group of linguistic cue words, the English word or phrase before the colon is corresponding English expression of Chinese cue words after the colon. The examples in the table are the translation of the original text.}
\end{table*}

\begin{table*}[htbp]
\setlength{\abovecaptionskip}{0pt}
\setlength{\belowcaptionskip}{0pt}
\begin{center}
\scalebox{0.70}{
\begin{tabular}{|l | p{5cm} | p{9cm} | rl |}
\hline \bf Group & \bf Cue Words & \bf Examples & \bf Percentage & \\ \hline
\hline
I: Prepositions  &  `{\bf for}': wei4, wei4le \newline \newline
`{\bf toward}': dui4, dui4yu2, dui4ci 
 & Most people feel angry \textbf{toward} incest wedding. &  143/511 =  \textbf{27.98\%} & \\ \hline

II: Conjunctions  & `{\bf because of}': yin1, yin1wei4, \newline you2yu2 \newline \newline
`{\bf so}': yu1shi4, suo3yi3, yin1er2 
 & He felt uneasy \textbf{because of} the deception from him. &  47/511 = \textbf{9.20\%} & \\ \hline

III: Light Verbs  & `{\bf to make}': rang4, ling4, shi3  & That Mr. Wu is rejected \textbf{made} him particularly frustrated. &  151/511 = \textbf{29.55\%} & \\ \hline

IV: Epistemic Markers  &  `{\bf to hear}': ting1, ting1dao4, \newline ting1shuo1 \newline \newline
`{\bf to see}': kan4, kan4dao4, \newline kan4jian4, jian4dao4, jian4, \newline kan4dao4, kan4jian4 \newline \newline
`{\bf to know}': zhi1dao4, de2zhi1, \newline de2xi1, huo4zhi1, huo4xi1, \newline fa1xian4, fa1jue2 
 & She is particularly excited \textbf{to see} these paintings. &  34/511 = \textbf{6.65\%} & \\ \hline

V: Passive  & `{\bf be adj by}': bei4
 & I \textbf{was} touched \textbf{by} their love.
 &  68/511 = \textbf{13.31\% }& \\ \hline

  &  &  & \bf 86.69\% & \\ \hline

\end{tabular}
}
\end{center}
\caption{ Linguistic cue words that characterize emotion cause at ``0'' position. For each group of linguistic cue words, the English word or phrase before the colon is corresponding English expression of Chinese cue words after the colon. The examples in the table are the translation of the original text.}
\end{table*}

\section{Re-evaluating ECE Models}
Previously, we had shown that a model without access to text can identify the emotion cause to an extent similar to the baselines because this model utilizes relative position information bias to help identify the location. This finding questions how much the state-of-the-art models are actually dependent on the non-position features they claimed to be using in their methods. 

The models below, which incorporate position information as features, are selected for performance re-evaluation after de-biasing dataset. The selected models are as follows:

\begin{itemize}
\item \textbf{PAE}: \citet{PAEDGL} proposed a Bi-LSTM based model with an attention mechanism to encode clauses along with a position feature;
\item \textbf{PAE-DGL}: \citet{PAEDGL} proposed a Bi-LSTM based model that jointly used text content, global label information and relative position altogether;
\item \textbf{RTHN} : \citet{RTHN} proposed a new hierarchical network architecture based on RNNs and Transformer. The lower layer is a word-level encoder consisting of multiple RNNs, and the upper clause-level encoder is a transformer to learn the correlation between multiple clauses. The relative position and global label are also encoded into transformer.
\end{itemize}

To control for the effect of position, we `de-bias' the dataset by separately extracting, as independent `datasets', the sets of instances for each position listed in the distribution in Table 2. We re-evaluated the models described above on these segmented datasets. Furthermore, to reduce the impact of position bias on performance of the models, we adjust the proportion of these four types of positions to a balanced ratio to form an adjusted dataset. The position distribution of the balanced dataset is shown in Table 8, which will be discussed in more detail below.

Table 6 shows the performance of each model on each new dataset. All the models make a great use of relative position information and thus perform extremely well on each single-position (e.g. dataset with only -1 clauses). However, the performance of each model decreases dramatically when it is re-valuated using the balanced dataset. In this scenario, relative position can't be used as an effective feature in finding emotion causes because of the equal distribution of the four position types. The results imply that the abilities of these deep neural network models to identify emotion causes and make predictions is far lower than previously perceived if position information do not contribute much useful information.

\begin{table}[ht]
\setlength{\abovecaptionskip}{0pt}
\setlength{\belowcaptionskip}{0pt}
\vspace{0cm}

\scalebox{0.78}{
\begin{tabular}{ |l|l|l|l|l| }
\hline
\bf Method & \bf Dataset & \bf P & \bf R &\bf F1 \\ \hline
\hline
\multirow{6}{*}{\bf PAE} & Dataset (Original) & 0.6897 & 0.6794  & \bf 0.6836 \\
& Dataset (Balanced) & 0.5511 & 0.3078  & \bf 0.3851 \\
& Dataset (only-1) & 1.0 & 0.9651  & 0.9819 \\
& Dataset (only0) & 1.0 & 0.9885  & 0.9941 \\
& Dataset (only1) & 1.0 & 0.9341  & 0.9651 \\
& Dataset (only-2) & 0.9941 & 0.8396  & 0.9050 \\
\hline
\hline
\multirow{6}{*}{\bf PAEDGL} & Dataset (Original) & 0.7484 & 0.6970  & \bf 0.7214 \\
& Dataset (Balanced) & 0.5526 & 0.3279  & \bf 0.4096 \\
& Dataset (only-1) & 1.0 & 0.9651  & 0.9819 \\
& Dataset (only0) & 1.0 & 0.9885  & 0.9941 \\
& Dataset (only1) & 0.9941 & 0.9341  & 0.9621 \\
& Dataset (only-2) & 0.995 & 0.8397  & 0.9061 \\
\hline
\hline
\multirow{6}{*}{\bf RTHN} & Dataset (Original) & 0.7642 & 0.7593  & \bf 0.7615 \\
& Dataset (Balanced) & 0.5467 & 0.5466  & \bf 0.5445 \\
& Dataset (only-1) & 0.9823 & 0.9542  & 0.9677 \\
& Dataset (only0) & 0.9882 & 0.9768  & 0.9825 \\
& Dataset (only1) & 0.9709 & 0.9333  & 0.9505 \\
& Dataset (only-2) & 0.9455 & 0.9144  & 0.9267 \\
\hline
\end{tabular}
}
\caption{Accuracy in dataset with balanced locations VS. biased locations.}
\end{table}

\begin{table}[ht]
\centering
\setlength{\abovecaptionskip}{5pt}
\setlength{\belowcaptionskip}{0pt}

\scalebox{0.66}{
\begin{tabular}{|c|c|c|c|c|c|}
\hline
\bf Position & \bf Benchmark & \bf Dataset1 & \bf Dataset2 &\bf Dataset3 & \bf Dataset4  \\ \hline
\hline
\bf -10   &    0.04\%&    0.05\%&       0.06\%&     0.08\%&      0.10\% \\
\bf -9   &    0.04\%&    0.05\%&       0.06\%&     0.08\%&      0.10\% \\
\bf -7   &    0.13\%&    0.15\%&       0.19\%&     0.26\%&      0.31\% \\
\bf -6   &    0.32\%&    0.36\%&       0.44\%&     0.61\%&      0.74\% \\
\bf -5   &    0.32\%&    0.36\%&       0.44\%&     0.61\%&      0.74\% \\
\bf -4   &    0.59\%&    0.68\%&       0.83\%&     1.05\%&      1.27\% \\
\bf -3   &    1.7\%&    1.79\%&       2.30\%&     2.82\%&      3.72\% \\
\bf -2   &    8.12\%&    8.79\%&       10.44\%&     13.32\%&      15.63\% \\
\bf -1 & \bf 54.45\% & \bf 48.65\%&    \bf 44.90\% &    \bf 36.71\%&   \bf 28.29\%  \\
\bf 0   &    23.58\%&    26.90\%&       25.62\%&     24.18\%&      24.04\% \\
\bf 1   &    7.47\%&    8.53\%&       10.24\%&     14.12\%&      17.12\% \\
\bf 2   &    2.21\%&    2.52\%&       3.07\%&     4.23\%&      5.10\% \\
\bf 3   &    0.50\%&    0.57\%&       0.70\%&     0.97\%&      1.17\% \\
\bf 4   &    0.18\%&    0.21\%&       0.25\%&     0.35\%&      0.42\% \\
\bf 5   &    0.09\%&    0.10\%&       0.18\%&     0.17\%&      0.21\% \\
\bf 7   &    0.04\%&    0.05\%&       0.06\%&     0.08\%&      0.10\% \\
\bf 8   &    0.09\%&    0.10\%&       0.12\%&     0.17\%&      0.21\% \\
\bf 12   &   0.04\%&    0.05\%&       0.06\%&     0.08\%&      0.10\% \\
\hline
\end{tabular}
}
\caption{Position distribution of original benchmark and bias-alleviated datasets. The proportion of dominant position (``-1'' emotion causes) is gradually decreased from 54.45\% to 28.29\%. }
\end{table}

To further prove the existence of bias, we re-evaluated the previously mentioned deep neural network models on four datasets generated by varying the proportion of cause clauses immediately preceding the emotion clause. The occurrence of ``-1'' and ``0'' clauses make up 78.03\% of the entire dataset, which conveys revealing position information of the large majority within the benchmark. Datasets with varying degrees of bias reduction are organized by continuously reducing the ratio of ``-1'', maintaining the ratio of ``0''and thereby automatically increasing the proportion of other positions within each dataset. As shown in Table 7, Benchmark is the original ECE dataset, and Dataset1 to Dataset4 are datasets with gradually decreasing degrees of ``-1'' reduction.

PAE, PAEDGL and RTHN are re-evaluated on these five groups of datasets as shown in Figure 2. The relationship between F1 measure and ratio of ``-1''clauses of these models shows a negative slope. Therefore, the performance of these models reduce steadily as “-1” clauses become less and less dominant in the dataset, which again demonstrates that these models are affected by position bias. In addition, PAE and PAEDGL's performance are more correlated with position information than that of RTHN.

\begin{figure}[htb] 
\setlength{\abovecaptionskip}{0pt}
\setlength{\belowcaptionskip}{0pt}
\centering
\includegraphics[width=3.0in]{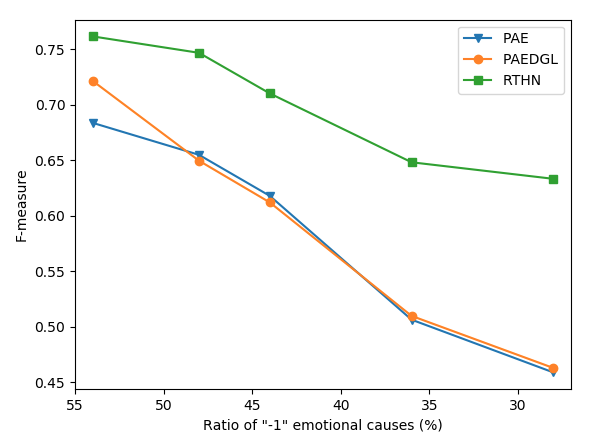} 
\caption{Accuracy in identifying emotion cause decreases with reduction of ``-1'' emotion causes.} 
\end{figure}

\section{Solution for Bias Reduction}
The bias in this benchmark is due to an overwhelming proportion of ``-1'' emotion causes. To alleviate the bias, we proposed the balanced dataset. By randomly selecting ``-1'', ``0'' and ``-2'' clauses and reducing each of the ratio to a similar level as ``1'' clauses while maintaining the same number of all other positions, we built a new balanced benchmark with 779 instances. The position distribution of balanced benchmark is shown in Table 8.

\begin{table}[ht]
\setlength{\abovecaptionskip}{0pt}
\setlength{\belowcaptionskip}{0pt}
\begin{center}
\scalebox{0.75}{
\begin{tabular}{|c|c|cl|}
\hline \bf Position & \bf Original & \bf Balanced & \\ \hline
\hline
Previous 10 Clauses & 0.04\% & 0.12\% &  \\
Previous 9 Clauses &  0.04\% & 0.12\% &  \\
Previous 7 Clauses &  0.13\% & 0.38\% & \\
Previous 6 Clauses &  0.32\% & 0.89\% & \\
Previous 5 Clauses &  0.32\% & 0.89\% & \\
Previous 4 Clauses &  0.59\% & 1.54\% & \\
Previous 3 Clauses & 1.70\% & 4.10\% & \\
\bf Previous 2 Clauses & \bf 8.12\% & \bf 18.87\% & \\
\bf Previous 1 Clauses & \bf 54.45\% & \bf 22.07\% &  \\
\bf In the same clauses & \bf 23.58\% & \bf 21.69\% &  \\
\bf Next 1 Clauses & \bf 7.47\% & \bf 20.41\% & \\
Next 2 Clauses & 2.21\% & 6.16\% & \\
Next 3 Clauses &  0.50\% & 1.41\% & \\
Next 4 Clauses &  0.18\% & 0.51\% & \\
Next 5 Clauses &  0.09\% & 0.25\% & \\
Next 7 Clauses &  0.04\% & 0.12\% & \\
Next 8 Clauses &  0.09\% & 0.25\% & \\
Next 12 Clauses &  0.04\% & 0.12\% & \\
\hline
\end{tabular}
}
\end{center}
\caption{ Position distribution of original and balanced benchmark. }
\end{table}

The previously mentioned models and our proposed random-selection approach are also re-evaluated on the balanced dataset, and Table 9 shows the performance of each model. The F-measure of the proposed random-selection approach decreases from 54.43\% to 24.04\%, which shows the efficacy of our solution in reducing position bias. Because other models' accuracy all decreased by at least 20\%, we proved that position no longer plays an as significant role as before.

\begin{table}[ht]
\setlength{\abovecaptionskip}{0pt}
\setlength{\belowcaptionskip}{0pt}
\begin{center}
\scalebox{0.90}{
\begin{tabular}{|l | l | l | rl |}
\hline \bf Method & \bf P & \bf R & \bf F1 & \\ \hline
\hline
Random & 0.2437 & 0.2373 & 0.2404 & \\ 
PAE & 0.5511 & 0.3078 & 0.3851 & \\ 
PAEDGL & 0.5526 & 0.3279 & 0.4096 & \\ 
RTHN & 0.5467 & 0.5466 & 0.5445 & \\ 
\hline
\end{tabular}
}
\end{center}
\caption{ Accuracy of emotion cause identification using balanced dataset. }
\end{table}

\section{Conclusion and Future Work}

Researchers usually pay more attention to developing better methods and improving accuracy results but often fail to verify whether the benchmark contains serious flaws that are able to compromise the dataset's intended test purpose. Some models use easy loopholes provided due to the shortcomings of the dataset to achieve high performance. More worryingly, neither the model nor the system designer may realize that the bias exists. As a result, others are misled into believing that it is the model's ability to learn deep semantic information that contributed to the rise of accuracy. However, it could be an irrelevant but significant bias within the benchmark that causes high performance. The non-interpretability of deep neural networks makes such verification even harder.

In this paper, we showed a simple random selection approach to solve the Emotion Cause Extraction task without observing the text, on a current benchmark, and showed that it achieved a similar results as the baselines. The high performance is purely due to the utilization of the relative position information of the emotion cause. Since position information alone without observing the text results in competitive F-measure, we uncovered a bias within the ECE benchmark. We studied the existence of the bias in the benchmark from a linguistic perspective, and concluded that one reason for this bias is that all instances belong to SINA news. The lack of diversity in genre and writing styles was therefore a contributing factor. We then proved that current high performance of deep neural network models that utilize location information is only evident in benchmarks that contain biases. The performance of these models decreased drastically after controlling for position in the benchmark. We proposed a method to improve the benchmark by debiasing it, and proved the efficacy of the debiased benchmark by comparing the accuracy result from our random approach to the results from three deep neural network models on the debiased benchmark.

Although our newly proposed balanced benchmark includes less position bias, the scale of the benchmark is also greatly reduced. In order to keep the same scale as the original benchmark, a potential future direction could be to rewrite the problematic benchmark, and achieve balance without affecting the scale. Literary text could also be extracted from more genres to ensure more diversity in linguistic expression across benchmark.

\bibliography{anthology,acl2020}
\bibliographystyle{acl_natbib}

\end{document}